\pdfoutput=1

\documentclass[11pt]{article}

\usepackage[final]{coling}

\usepackage{times}
\usepackage{latexsym}
\usepackage{graphicx}
\usepackage{comment}
\usepackage{xcolor}
\usepackage{multirow}
\usepackage{graphicx}
\usepackage{algorithm}
\usepackage{algorithmic}
\usepackage{verbatimbox}
\usepackage{tcolorbox}

\usepackage[T1]{fontenc}

\usepackage[utf8]{inputenc}

\usepackage{microtype}

\usepackage{inconsolata}

\usepackage{graphicx}

\title{Multi-Document Grounded Multi-Turn Synthetic Dialog Generation}

\author{First Author \\
  Affiliation / Address line 1 \\
  Affiliation / Address line 2 \\
  Affiliation / Address line 3 \\
  \texttt{email@domain} \\\And
  Second Author \\
  Affiliation / Address line 1 \\
  Affiliation / Address line 2 \\
  Affiliation / Address line 3 \\
  \texttt{email@domain} \\}

\author{
Young-Suk Lee$^*$, Chulaka Gunasekara\thanks{Primary authors with equal contributions}, Danish Contractor \\
\bf{Ramón Fernandez Astudillo, Radu Florian} \\
 \texttt{\{ysuklee,raduf\}@us.ibm.com} \\
 \texttt{
 \{Chulaka.Gunasekara,Danish.Contractor,ramon.astudillo\}@ibm.com} \\ 
 IBM Research AI
  } 

\begin{document}
\maketitle
\begin{abstract}
We introduce a technique for multi-document-grounded multi-turn synthetic dialog generation that incorporates three main ideas. First, we control the overall dialog flow using taxonomy-driven user queries that are generated with Chain-of-Thought (CoT) prompting. Second, we support the generation of multi-document-grounded dialogs by mimicking real-world use of retrievers to update the grounding documents after every user-turn in the dialog. Third, we apply LLM-as-a-Judge to filter out queries with incorrect answers.
Human evaluation of the synthetic dialog data suggests that the data is diverse, coherent, and includes mostly correct answers. 
Both human and automatic evaluations of answerable queries indicate that models fine-tuned on synthetic dialogs consistently out-perform those fine-tuned on existing human generated training data across four publicly available multi-turn document grounded benchmark test sets.  
\end{abstract}

\section{Introduction}

\begin{figure*}[t]
    \centering
    \includegraphics[width=\textwidth]{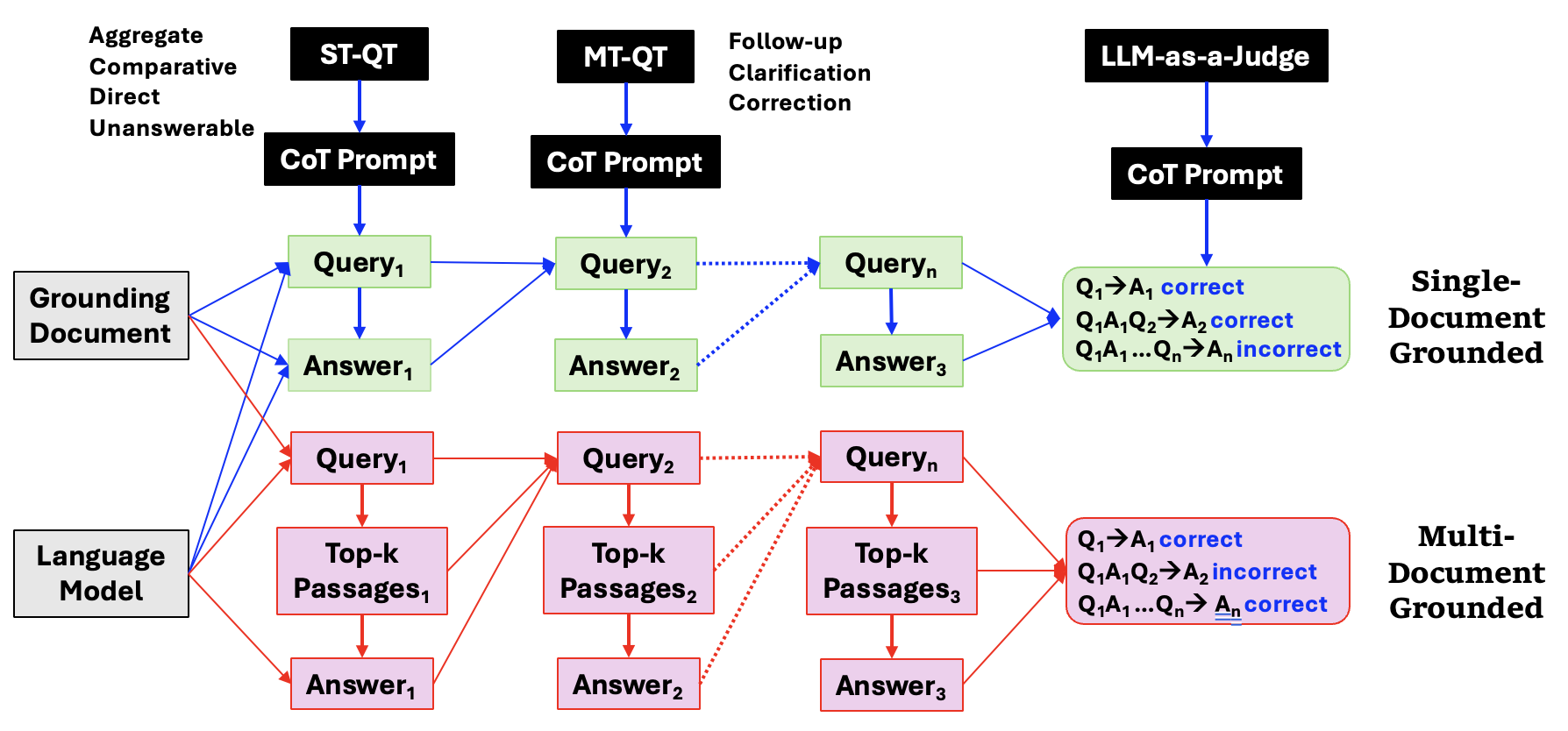}
    \vspace{-6mm}
    \caption{Overview of document grounded multi-turn synthetic dialog generation pipeline. We distinguish two types of dialog style, single-document grounded (light green color boxes) and multi-document grounded (retrieval augmented generation, pink color boxes). Both styles share the same starting-turn query taxonomy (ST-QT), CoT prompt for the initial query (Query$_{1}$) generation, multi-turn query taxonomy (MT-QT) and the CoT prompt for second turn query (Query$_{2}$) generation. User queries and agent answers are generated by an LLM.
    Given the initial query generated from the same single document, single-document grounded dialog generation proceeds according to Algorithm~\ref{alg:mrc-dialog-algorithm} and multi-document-grounded, according to Algorithm~\ref{alg:rag-dialog-algorithm} in \S{\ref{sec:dialog-flow}}. After generating multi-turn dialogs, we apply LLM-as-a-Judge to filter out queries with incorrect answers turn-by-turn. We use \textsc{Mixtral-8x7b-instruct} as our language model for both data generation and LLM-as-a-Judge.}
    \label{fig:sdg-pipeline}
\end{figure*}

As instruction-tuned language models have proven highly effective to generalize to new tasks, \cite{flan-t5-2022, flan2021,instructgpt, naturalinstructions, supernaturalinstructions}, there has been growing interest to acquire synthetic data sets generated from pre-trained language models with a minimal or no human supervision, \cite{unnatural2022, self-instruct2023, baize2023, ensemble-instruct2023}. 
While there has been an exploration of synthetic data generation for persona-grounded dialog generation~\cite{persona1,persona2}, topic-grounded dialog generation~\cite{wsdm2024}, summary-grounded dialog generation~\cite{gunasekara2021summary} and conversation generation from dialog scenarios ~\cite{mohapatra-etal-2021-simulated-chats}, the study of methods for generating multi-document grounded multi-turn dialogs by mimicking real-world deployments with retrievers has been limited ~\cite{persona2, wsdm2024}. Generating a realistic multi-turn document grounded dialog can be challenging as one needs to ensure that the (i) the questions are diverse (ii) conversation flows naturally - i.e, it is coherent and the follow-up questions are more than just a collection of question-answer pairs with co-references (iii) the responses are faithful to the documents in the retrieved set and not generated from model parameters.     

In this paper, we address these challenges and present a technique for generating high quality document grounded multi-turn synthetic dialogs. Our data generation flow incorporates two novel ideas - first, we control the dialog flow according to a question-taxonomy with chain-of-thought (CoT) prompting for query generation. Second, we generate dialogs grounded on both single and multiple retrieved documents. We also apply self-consistency \cite{wang2022self} and an LLM-as-judge \cite{mtbench2023} to ensure the answers are correct. 

For multi-document grounded dialog generation, 
user queries and agent answers are based on top-$k$ retrieved passages. In particular, we generate an initial user query from a single document source and generate the agent answer from top-k passages retrieved on the initial user query. Subsequent user queries and all agent answers are grounded on the retrieved passages and dialog history. We use a series of carefully designed prompts to ensure generated agent answers continue to remain meaningful in the presence of retrieved passages, often noisier than human generated documents. 
We use \textsc{Mixtral-8x7b-instruct} as our language model for both data generation and LLM-as-a-Judge throughout this paper. However, the proposed synthetic data generation framework is not tied to any specific language model.
An overview of the proposed technique is shown in Figure~\ref{fig:sdg-pipeline}.

To assess the quality of the synthetic data, we carry out extensive human evaluations on $294$ dialogs comprising $712$ user-agent turns. 
Our studies indicate that the data generated by our pipeline is diverse, coherent, and includes largely correct answers, across different dialog turn-lengths.  
Further, we conduct rigorous experiments by training two instruction-tuned models \textsc{merlinite-7b}\footnote{https://huggingface.co/instructlab/merlinite-7b-lab}~\cite{sudalairaj2024lab} and \textsc{llama-2-13b-chat}\footnote{https://huggingface.co/meta-llama/Llama-2-13b-chat-hf}~\cite{touvron2023llama} using our synthetic data and study the performance on four publicly available multi-turn human-generated benchmark data sets,~CoQA~\cite{coqa2019}, MultiDoc2Dial ~\cite{multidoc2dial2021}, QuAC~\cite{quac2018} and OR-QuAC ~\cite{orquac2020}. 
Our experiments indicate that the two language models fine-tuned on our synthetic data consistently out-perform models fine-tuned on existing (human-generated) training data of comparable sizes using both automated metrics and human evaluations for answerable queries. We also demonstrate how the use of retriever-augmented synthetic data is necessary for doing well on end-tasks that simulate real-world RAG usage.   

In summary, our paper makes the following contributions: (i) To the best of our knowledge, we present the first data generation pipeline that mimics real-world RAG deployments for content-grounded dialogs, (ii) We ensure higher diversity in the data by relying on question types from a taxonomy rather than solely relying on the LM to generate queries. (iii) We conduct comprehensive human studies assessing the quality the generated dialogs 
(iv) We demonstrate the effectiveness of the synthetic data over human generated benchmark data for answerable queries
by conducting rigorous automated and human evaluations. (v) Lastly, we will release all our code and synthetic data.    


\section{Multi-turn Dialog Generation}
\label{sec:mtgen}

Our multi-turn dialog generation consists of four components: (1) Two question taxonomies  - one designed for the first turn of the dialog (ST-QT) and the other for subsequent turns (MT-QT). 
(2) Chain-of-Thought (CoT) prompting to generate desired query types according to question taxonomy, (3) Two separate synthetic data generation pipelines for single-document and multi-document grounded dialogs. (4) LLM-as-a-Judge to filter out dialogs containing incorrect answers for any given queries.

\subsection{Question Taxonomy}
\label{sec:question-taxonomy}

To ensure that generated dialogs are diverse 
we devise a question type taxonomy. This taxonomy is a subset of contemporaneous ideas presented in ~\cite{yang2024crag} and covers the most critical query types found in RAG datasets. We adopt the following question taxonomy for all initial user queries:

\noindent\textbf{Direct}: Free form questions whose answer can be directly extracted from the grounding document\\
\noindent\textbf{Comparative}: Questions that request comparisons between different entities and/or concepts\\
\noindent\textbf{Aggregate}: Questions that require synthesis of information from multiple documents or paragraphs\\
\noindent\textbf{Unanswerable}: Questions that request information   unavailable in the given documents

For subsequent user turns, we use the following:

\noindent\textbf{Follow-up}: Questions that build on previous responses, typically with elliptical phrases such as \textit{what about, how about, etc.}\\
\noindent\textbf{Clarification}: Questions that seek to resolve ambiguities, as in \textit{you mean, does that mean}\\
\noindent\textbf{Correction}: Questions that aim to rectify errors or misunderstandings from earlier conversations, as in \textit{no, that's not what I meant}\\

\subsection{Query generation with CoT Prompting}
\label{sec:cot-prompting}

To ensure that the generated queries 
conform the pre-specified question taxonomy, we incorporate question-type specific CoT prompts that instruct an LLM to reason through the grounding document.

While most previous work adopts CoT prompts for accurate answer generation in reasoning tasks \cite{cot-neurips2022, cot-neurips2022b, react-iclr2023}, we found it effective for steering an LM to generate queries of desired properties. We use a CoT prompt in zero-shot setting for initial turns and few-shot setting for subsequent turns. In our experiments, we found that using $3$ in-context learning (ICL) examples of a question type (ex: comparative questions) without CoT prompt resulted in only 5\% of the intended query type, 
whereas CoT prompt for a specific query type without any ICL examples resulted in more than 90\% of the generated queries adhering to the intended query type.

Our data generation framework is flexible enough to incorporate any changes to the taxonomy, as a user could create new prompts for additional question types, and run the rest of the data generation pipeline.

\subsection{Response generation with CoT Prompting}
\label{sec:cot-prompting-resp}

For generating the agent answers, we utilize another CoT prompt. This CoT prompt includes a reasoning step to explain how the model looks for the answer in the available documents, and then generates the answer grounded on the documents and generated reasoning. To ensure that the model generates consistent (w.r.t. the generated reasoning) and faithful (w.r.t. the documents), the CoT prompt instructs the LLM to perform a consistency check (inspired by~\citet{wang2022self}), and extract evidences from the documents. 
This dual use of CoT for both query and answer generation 
ensures that the interactions are both rich and logically consistent. We use greedy decoding for each user and agent turn generation and top-k sampling (k=50)~\cite{fan2018hierarchical} for all other steps. All of the CoT prompt templates used in this study are available in Figures~\ref{fig:direct-prompt}-\ref{fig:correction-prompt} (Appendix).

\subsection{Dialog Generation}
\label{sec:dialog-flow}

Once the initial user queries are generated according to the question taxonomy in \S{\ref{sec:question-taxonomy},
initial agent answers and all subsequent user/agent turns are generated by two separate pipelines - single-document grounded pipeline in \S{\ref{sec:mrc-style}} and multi-document-grounded pipeline in \S{\ref{sec:rag-style}}. 
\subsubsection{Single-document grounded dialog}
\label{sec:mrc-style}

\begin{algorithm}[ht]
\small
\caption{Single-document Grounded Dialog Generation}
\label{alg:mrc-dialog-algorithm}
\textbf{Input}: Grounding document $doc_{g}$ \\
\textbf{Output}: $dialog_{n}$ 

\begin{algorithmic}[1]
\STATE $history \leftarrow \phi$  


\FOR {$i$ in $1~...~n$}
\STATE $q_{i} \leftarrow LLM(history_{i-1}, doc_{g}$)
\STATE $a_{i} \leftarrow LLM(q_{i}, history_{i-1}, doc_{g}$)
\STATE $history_{i} \leftarrow 
 history_{i-1} \cup q_{i} \cup a_{i}$

\ENDFOR
\STATE $dialog_{n} \leftarrow history_{n}$
\STATE return $dialog_{n}$
\end{algorithmic}
\end{algorithm}

In this setup all user and agent turns are grounded on the same gold document. In Algorithm~\ref{alg:mrc-dialog-algorithm}, each query $q_{i}$ (line 3) is generated by an LM on the basis of dialog history and the grounding document $doc_{g}$. Given query, dialog history and $doc_{g}$, the LM generates an answer $a_{i}$ (line 4). The query and the answer generated at each turn is added to the dialog history $history_{i}$ (line 5). Finally, $history_{n}$ is returned as the output $dialog_{n}$ (lines 7-8).

\subsubsection{Multi-document grounded dialog}
\label{sec:rag-style}

\begin{algorithm}[ht]
\small
\caption{Multi-document Grounded Dialog Generation}
\label{alg:rag-dialog-algorithm}
\textbf{Input}: Initial user query $q_{1}$ from Algorithm~\ref{alg:mrc-dialog-algorithm} \\
\textbf{Output}: $dialog_{n}$ \& $passages_{m}$ \\
\begin{algorithmic}[1]
\STATE $history \leftarrow \phi$ ; $passage_{m} \leftarrow \phi$

\FOR{ i = 1}
\STATE $passage_{1,k} \leftarrow retriever(q_{1})$ \\
\STATE $passage_{m} \leftarrow passage_{1,k}$ \\
\STATE $a_{1}\leftarrow LLM(q_{1},passage_{m})$ \\
\STATE $history_{1} \leftarrow q_{1}~\cup~a_{1}$ \\
\ENDFOR

\FOR {$i$ in $2~...~n$}
\STATE $q_{i}\leftarrow LLM(history_{i-1}, passage_{m})$
\STATE $passage_{i,k} \leftarrow retriever(q_{i}, history_{i-1})$
  \FOR {$j$ in $1~...~k$}
    \IF {$passage_{i,j} \notin passage_{m}$}
      \STATE $passage_{m} \leftarrow passage_{m} ~\cup~passage_{i,j}$
    \ENDIF
  \ENDFOR
\STATE $a_{i} = LLM(q_{i}, history_{i-1}, passage_{m})$
\STATE $history_{i} \leftarrow 
 history_{i-1} \cup q_{i} \cup a_{i}$

\ENDFOR
\STATE $dialog_{n} \leftarrow history_{n}$
\STATE return $dialog_{n}~\&~passage_{m}$
\end{algorithmic}
\end{algorithm}

\noindent
For multi-document grounded dialog, we integrate a retriever that dynamically selects top-$k$ relevant passages from a pre-constructed document index. In Algorithm~\ref{alg:rag-dialog-algorithm}, a dialog starts with the initial query generated by Algorithm~\ref{alg:mrc-dialog-algorithm}. At turn 1, the retriever retrieves top-k passages $passage_{1,k}$ given the seed query $q_{1}$ and store them as $passage_{m}$ (lines 3-4). Given the query and grounding passages, an LM generates an answer $a_{1}$ (line 5) and $q_{1}$ and $a_{1}$ get stored in the dialog history $history_{1}$ (line 6). 

For subsequent turns, the LM generates a query given dialog history and retrieved passages (line 9). The retriever retrieves top-k passages given query and dialog history.  Each retrieved passage is added to $passage_{m}$ if and only if the passage is new (lines 11-15). The LM generates an answer given query, dialog history and unique passages retrieved so far, $passage_{m}$ (line 16). Finally, the dialog $history_{n}$ is returned as the output $dialog_{n}$ along with $passage_{m}$, a collection of unique passages retrieved during the session.

\subsubsection{Retriever}
 Each document is split into passages with maximum 512 tokens with 100 overlapping tokens between passages for indexing. We indexed the 11,377,951 passages from \cite{orquac2020}\footnote{https://github.com/prdwb/orconvqa-release} to compare the performance of models fine-tuned on human generated benchmark training set and synthetic dialogs generated from the corresponding retrieved passages.
 As our retriever, we use ELSER (Elastic Learned Sparse EncodeR)\footnote{https://www.elastic.co/guide/en/machine-learning/current/ml-nlp-elser.html} with all-MiniLM-L6-v2\footnote{https://huggingface.co/sentence-transformers/all-MiniLM-L6-v2} sentence embeddings. 

\subsection{LLM-as-a-Judge for Correctness}

We incorporate an LLM-as-a-Judge module to filter out queries with incorrect answers. For each generated dialog, we first extract all context-response pairs, consisting of dialog history plus the current query and its answer, as shown in Figure~\ref{fig:sdg-pipeline}. We then ask an LLM to judge whether the current answer is correct or not given the dialog history and the current query. We filter out all context-response pairs which are judged to be incorrect. We include the CoT prompt of the LLM-as-a-Judge in Figure~\ref{fig:llm-judge} (Appendix).

\subsection{Data Quality Evaluation}
\label{sec:sdg-evaluation}
\begin{table*}
\small
\centering
\begin{tabular}{ccccccc}\\\hline
\textbf{Dialog Length}  & \textbf{\#Dialog} & \textbf{\%correct} & \textbf{\%answerable} & \textbf{\%plausible} & \textbf{\%diverse} & \textbf{\%coherent} \\\hline
1 turn & 100 & 75.6 & 85.2 & 93.4 & na &  na\\
2 turns & 60  & 73.1 & 87.4 & 98.9 & 98.9 & 100.0 \\
3 turns & 60  & 62.1 & 78.1 & 96.2 &92.9 & 95.0 \\
4 turns & 58  & 76.3 & 88.4 & 98.0 & 98.0 & 95.0 \\
5 turns & 16  & 88.8 & 93.8 & 99.4 & 90.6 & 93.8 \\\hline
Overall  & 294 & 73.1 & 85.2 & 96.7 & 95.6 & 96.7 \\\hline
\end{tabular}
\caption{Human evaluation of synthetic data quality according to dialog length. Length 5 indicates 5 user queries plus 5 agent answers. \%diverse and \%coherent are applicable only to dialogs with 2 turns or more. Each sample has been evaluated by 2 or more annotators.}
\label{tab:sdg-human-evaluation-results}
\end{table*}
To assess the quality of the synthetic data \cite{fed2020, nlp4convai2022}, we carry out human evaluations on $294$ dialogs comprising $712$ user-agent turns. We use the following criteria for evaluation: 
{\em plausibility} and {\em answerability} for user queries, {\em correctness} for agent answers, {\em diversity} and {\em coherency} for the overall dialog. Each criterion is described in Table~\ref{tab:sdg-evaluation-criteria} (Appendix). We trained $7$ professional annotators with $3$ or more years of work experience with the instructions in Figure~\ref{fig:sdg-evaluation-instruction} (Appendix).

\begin{table}[ht]
\centering
\small
\vspace{-0.2cm}
\begin{tabular}{llll}\\\hline
           &  \textbf{E} & \textbf{H} & \textbf{R} \\\hline
\textbf{C} & 90 (45/50) & 65 (71/110) & 77 (46/60) \\
\textbf{E} &         & 74 (37/50) & 72 (43/60) \\
\textbf{H} &         &       & 42 (50/120) \\\hline
\end{tabular}
\caption{Inter-annotator agreement (IAA) w.r.t the answer correctness across 4 annotators, C, E, H and R. Included in parentheses are raw numbers.}
\label{tab:IAA}
\end{table}
\noindent{\bf Results:} The results are summarized in Table~\ref{tab:sdg-human-evaluation-results}. 
The synthetic dialogs exhibit a very high degree of diversity and coherency and most queries are considered highly plausible. Answer correctness is $73.1$\%. The ratio between answerable and unanswerable queries is $85.2$ vs.~$14.8$. 
Inter-annotator agreements range from 42\% to 90\% with around 70\% on average, as shown in Table~\ref{tab:IAA}. 
Table~\ref{tab:correlation-between-answerability-correctness} indicates an apparent correlation between query answerability and answer correctness. Answerable queries are more likely to lead to correct answers than unanswerable queries.

\begin{table}
\small
\centering
\vspace{-0.2cm}
\begin{tabular}{cccc}\\\hline

\textbf{Turn Depth}  & \textbf{\#Turns} & \textbf{\%correct} & \textbf{\%answerable}  \\\hline
1 & 100 & 76.8 & 87.5 \\
2  & 120  & 68.2 & 83.7 \\
3  & 180  & 66.4 & 79.3  \\
4  & 232  & 77.9 & 87.8 \\
5  & 80  & 84.4 & 93.8  \\\hline
Overall  & 712 & 73.1 & 85.2  \\\hline
\end{tabular}
\caption{Correlation between query answerability and answer correctness. Out of the 1270 annotations with answerable queries, 1062 (83.6\%) had correct answers.}
\label{tab:correlation-between-answerability-correctness}
\end{table}

\section{Experimental Results}
\label{sec:experiments}
We answer the following questions through our experiments:
(1) What is the difference in performance when using generated synthetic data in place of human-generated data for training content-grounded conversational models? 
(2) Is multi-document grounded data generation useful for tasks evaluated in multi-document settings, e.g. RAG?

\noindent{\bf Models:} We demonstrate the efficacy of our proposed multi-turn synthetic dialog generation technique by comparing the performance of models fine-tuned on the synthetic data and those fine-tuned on human generated benchmark data. We fine-tune two open-sourced LLMs, i.e.~\textsc{merlinite-7b} and \textsc{llama-2-13b-chat}. 

\noindent{\bf Benchmarks:} We conduct experiments on four open-sourced multi-turn benchmark datasets:  MultiDoc2Dial,
CoQA,
QuAC
and OR-QuAC.
We report all results on the test portion of the benchmarks for MultiDoc2Dial and OR-QuAC. For CoQA and QuAC for which test sets are not available, we use development set as the test set and keep a subset of the training data as validation set. For CoQA, we use 2 types of reference answers, one short answer, the official reference answer and the other, rationale from which short answers are derived. The train, dev and test split sizes of each dataset is shown in Table~\ref{tab:data_sizes} (Appendix).

\noindent{\bf Synthetic Data: }We obtain synthetic counterparts of the human-generated benchmark training sets. For CoQA, MultiDoc2Dial, QuAC data sets, the grounding documents of which consist of single documents, we generate the synthetic dialogs according to Algorithm~\ref{alg:mrc-dialog-algorithm}. For OR-QuAC, the grounding documents of which consists of top-k retrieved passages, we generate the synthetic data according to Algorithm~\ref{alg:rag-dialog-algorithm}. Table~\ref{tab:sdg_stats} (Appendix) summarizes the statistics of the generated synthetic dialogs.

\begin{table*}
\footnotesize
\centering
\begin{tabular}{llcccc}\hline
\textbf{Benchmarks/Models} & \textbf{TrainSet} & \textbf{F1} & \textbf{RougeL} & \textbf{Bert-Recall} & \textbf{Recall} \\\hline
\textbf{QuAC} (665)           \\\hline
Mixtral-8X7B-Instruct-v0.1 & zero-shot & 0.295 & 0.292  & 0.382  & 0.525 \\\hline
               & zero-shot & 0.325  & 0.324 & 0.393 & \textbf{0.513} \\
LlaMA2-13B-Chat & human benchmark  & 0.366 & 0.379 & 0.299 & 0.373 \\
 &   synthetic  & 0.393 & \textbf{0.395} & 0.402 & 0.479  \\
 & synthetic-filtered & \textbf{0.394} & 0.393 & \textbf{0.418} & 0.504 
 \\\hline
        & zero-shot & 0.294 & 0.290 & 0.374 & 0.509 \\
Merlinite-7B & human benchmark & 0.367 & 0.382 & 0.282 & 0.368 \\
   &  synthetic & 0.394 & 0.393 & 0.431 & 0.509 \\
   & synthetic-filtered & \textbf{0.397} & \textbf{0.393} & \textbf{0.449} & \textbf{0.537} \\
\hline
\textbf{OR-QuAC} (4421)          \\\hline
Mixtral-8X7B-Instruct-v0.1 & zero-shot & 0.191  & 0.183   & 0.256 & \textbf{\textit{0.367}}  \\\hline
                & zero-shot  & 0.122 & 0.118   &  0.132   & 0.244     \\
LlaMA2-13B-Chat & human benchmark  & 0.145 & 0.149 & 0.133 & 0.153 \\
 & synthetic  & 0.153 & 0.154 & 0.163 & 0.187 \\
 & synthetic-filtered & \textbf{0.225} & \textbf{0.223} & \textbf{0.263} & \textbf{0.295} \\
\hline
             & zero-shot & 0.168  & 0.166 & 0.186  & 0.262  \\
Merlinite-7B & human benchmark & 0.142  & 0.145 & 0.130  & 0.149 \\
  & synthetic & 0.199 & 0.197 & 0.233 & 0.265 \\
 & synthetic-filtered & \textbf{0.225} & \textbf{0.221} & \textbf{0.278} & \textbf{0.323} \\
\hline
\textbf{MD2D} (6178)           \\\hline
Mixtral-8X7B-Instruct-v0.1 & zero-shot & 0.226 & 0.258   & 0.301   & 0.396 \\\hline
                & zero-shot &  0.227  &  0.266    & 0.296     & 0.399     \\
LlaMA2-13B-Chat & human benchmark  & 0.406 &	0.389 &	0.376 &	0.407 \\
   &  synthetic  & 0.429 &	0.407 & 0.412 &	0.493 \\
   &  synthetic-filtered  & \textbf{0.431} & \textbf{0.411} & \textbf{0.431} &	\textbf{0.501} \\
\hline
              & zero-shot  & 0.218 & 0.200 & 0.288  & 0.386  \\
Merlinite-7B  & human benchmark  & 0.409 & 0.392 & 0.394 &	0.430  \\
   & synthetic  & 0.439 & 0.421 & 0.415 & 0.450 \\
   & synthetic-filtered  & \textbf{0.442} & \textbf{0.429} & \textbf{0.416} & \textbf{0.454} \\
\hline
\textbf{CoQA} w/ rationale (7918)           \\\hline
Mixtral-8X7B-Instruct-v0.1 & zero-shot & 0.209   & 0.197  &  0.294  &  \textbf{\textit{0.560}}   \\\hline
               & zero-shot &  0.236    &    0.227  & 0.235     & \textbf{0.470}  \\
LlaMA2-13B-Chat & human benchmark  & 0.378 &	0.379 &	0.088 &	0.320	 \\
 & synthetic  & 0.430 &	0.430 &	0.226 &	0.428   \\
 & synthetic-filtered  & \textbf{0.441} & \textbf{0.439} & \textbf{0.229} &	0.467   \\
\hline
            & zero-shot & 0.264     & 0.257      &  0.246        &  0.478   \\
Merlinite-7B & human benchmark  &  0.381 &	0.384 &	0.092 &	0.323  \\
  &  synthetic  &  0.445 & 0.441 &	0.311 & 0.499  \\
 &  synthetic-filtered  &  \textbf{0.449} &	\textbf{0.441} &	\textbf{0.316} & \textbf{0.511}  \\
\hline
\textbf{CoQA} w/ short answer (7918) \\\hline
Mixtral-8X7B-Instruct-v0.1 & zero-shot & 0.154   &  0.153   &  0.348    &  0.852  \\\hline
                & zero-shot & 0.256   & 0.277     & 0.497      &  \textbf{0.848}  \\
LlaMA2-13B-Chat & benchmark  & \textbf{0.811} & \textbf{0.807} & \textbf{0.734} &	0.813	 \\
  & synthetic  & 0.667 &	0.663 &	0.629 &	0.834 \\
  & synthetic-filtered  & 0.669 &	0.661 &	0.628 &	0.836\\
\hline
            & zero-shot &   0.274    & 0.269     &  0.438     & 0.807      \\
Merlinite-7B  & benchmark  &  \textbf{0.819} & \textbf{0.816} & \textbf{0.741} &	0.821  \\
  &  synthetic  &  0.532 &	0.528 &	0.525 &	\textbf{0.854}  \\
  &  synthetic-filtered  &  0.546 &	0.535 &	0.523 &	0.853  \\
\hline
\end{tabular}
\caption{Testset performances on \textbf{answerable queries}. Comparisons are among the models fine-tuned on \textit{human annotated benchmark} training data vs. \textit{synthetic} (without LLM-as-a-Judge) vs. \textit{synthetic-filtered} (with LLM-as-a-Judge) data. The highest scores in each condition are bold-faced. Models fine-tuned only on the synthetic data out-perform those fine-tuned on the human benchmark data. 
The only exception is the CoQA data set scored with short answer as reference, for which the models trained on the synthetic data under-perform those trained on the human benchmark data except for Recall. However, we show in \S{\ref{sec:human-evaluation}} that humans overwhelmingly prefer outputs generated by the models fine-tuned on the synthetic data to those by the models trained on the human benchmark data. \textit{Synthetic-filtered} is particularly effective for OR-QuAC, indicating that synthetic dialogs generated from retrieved passages are noisier than those generated from human generated single document and therefore, post-filtering improves the model accuracy by a significant margin.}
\label{tab:only-answerable-results}
\end{table*}

\noindent{\bf Model Training:} We (Q)LoRA fine-tune \cite{lora2021, dettmers2023qlora} all models
on three types of training data: (1) original human-generated benchmark training data, (2) synthetic data without the LLM-as-a-Judge filter (and with unanswerable queries), (3) synthetic data with the LLM-as-a-Judge filter (and without unanswerable queries). 
We fine-tune all models 
on 4 A100/80GB GPUs. We set the quantization method to `fp4', lora rank=8, lora alpha=32, batch size=16. We use `DeepSpeedFusedAdam' as the optimizer with peak learning rate = 1-e5, weight decay = 0.1, betas = [0.9, 0.95] and eps = 1-e10. We train each model up to 5 epochs and select the checkpoint with the highest automatic scores on the validation set as the best model.

\subsection{Automatic Evaluations}


We evaluate the model output qualities with 4 automatic metrics: (1) F1, standard metric used in \citet{quac2018}, \citet{coqa2019}, \citet{multidoc2dial2021}, (2) RougeL, (3) Bert-Recall, (4) Recall, the metric reported to achieve the highest correlation with human evaluation for answer correctness of instruction following models in question answering  \cite{tacl2023correctness}. 
We study the model performance in four settings: (i) zero-shot, (ii) human-generated benchmark training data (human benchmark), (iii) synthetic data with unanswerable queries and without the LLM-as-a-Judge filter (synthetic), (iv) synthetic data without unanswerable queries and with the LLM-as-a-Judge filter (synthetic-filtered). We report the performance of the teacher model under zero-shot settings. 


\begin{table*}[ht]
\small
\centering
\begin{tabular}{lcccccccc} \hline
 & \multicolumn{2}{c}{CoQA} & \multicolumn{2}{c}{MD2D} & \multicolumn{2}{c}{QuAC} & \multicolumn{2}{c}{OR-QuAC} \\\hline
models & LlaMA & Merlinite & LlaMA & Merlinite & LlaMA & Merlinite & LlaMA & Merlinite \\\hline
reference & 2.5 / 9.1 &  2.5 / 9.1 & 15.6 & 15.6 &  12.6 & 12.6 & 13.2 & 13.2  \\\hline
human benchmark & 2.3 & 2.3 & 13.8   & 15.5 & 11.0 & 10.2 & 9.5 & 9.4 \\
synthetic & 6.1 & 9.1 & 31.2 & 35.8 & 18.0 & 20.9 & 13.4 & 19.7 \\\hline
\end{tabular}
\caption{Output length distribution of fine-tuned models on the 4 benchmark test sets. 
For CoQA, we have 2 types of references, short answer with 2.5 tokens and rationale with 9.1 tokens on average.}  
\label{tab:output_length_distribution}
\end{table*}

Table~\ref{tab:only-answerable-results} shows the results on the subset of the test sets with answerable queries.  
Across all benchmarks, QuAC, OR-QuAC and MultiDoc2Dial and CoQA scored with rationale as reference, models fine-tuned on the synthetic data out-perform those fine-tuned on human generated benchmark data, corroborating the efficacy of our proposed technique. While models fine-tuned on synthetic-filtered mostly out-perform those fine-tuned on synthetic, synthetic-filtered is particularly effective for OR-QuAC, indicating that synthetic dialogs generated from retrieved passages are noisier than those generated from human generated documents and therefore, LLM-as-a-Judge filter improves the model accuracy by a significant margin. 

\noindent{\bf Unanswerable queries: } Unanswerable queries pose challenges for both evaluation and multi-document grouned generation. For evaluation, there is a complete mismatch between the reference answer and the LM-generated answer for QuAC and OR-QuAC. While human benchmark answers are categorical, as in \textit{I do not have the information}, language model responses are highly varied, as in \textit{Query: How much did Deloitte Touche Tohmatsu contribute to Fossella’s campaign in 2006 and 2008? Answer: Deloitte Touche Tohmatsu was the second largest campaign contributor to Fossella in the 2006 campaign cycle and among the largest contributor in the 2008 campaign cycle. However, the document does not provide specific amounts for these contributions}.  For multi-document grounded generation, there is no straightforward way of applying unanswerable query taxonomy to document retrieval,\footnote{An option is to apply an answerability judge to the generated queries on the basis of the retrieved document post-hoc.} which always tries to retrieve relevant documents to a given user query. Therefore, we leave evaluation and generation of unanswerable queries in a multi-document grounded dialog generation setup to future work. For completeness, we report the performance on unanswerable queries in Table~\ref{tab:only-unanswerable-results} (Appendix).

\noindent \textbf{CoQA: Short Answer vs. Rationale:}
In Table~\ref{tab:only-answerable-results}, CoQA scored with short answer behaves differently from the other three benchmark test sets and CoQA scored with rationale. 
We hypothesize that this may be attributed to the exceptional shortness of the CoQA short answers.
Average length of CoQA short answer is 2.5 tokens compared with 13-16 tokens in other test sets or 9.1 tokens in rationale, as shown in Table~\ref{tab:output_length_distribution}. The lower F1 and RougeL with short answer may be a consequence of the fact that the long but correct hypotheses are unfairly penalized for precision. Therefore, we turn to human evaluations to determine which model output is better aligned with human preferences in \S{\ref{sec:human-evaluation}}.

\subsection{Human Evaluation}
\label{sec:human-evaluation}
\begin{figure}
    \centering
    \includegraphics[scale=0.52]{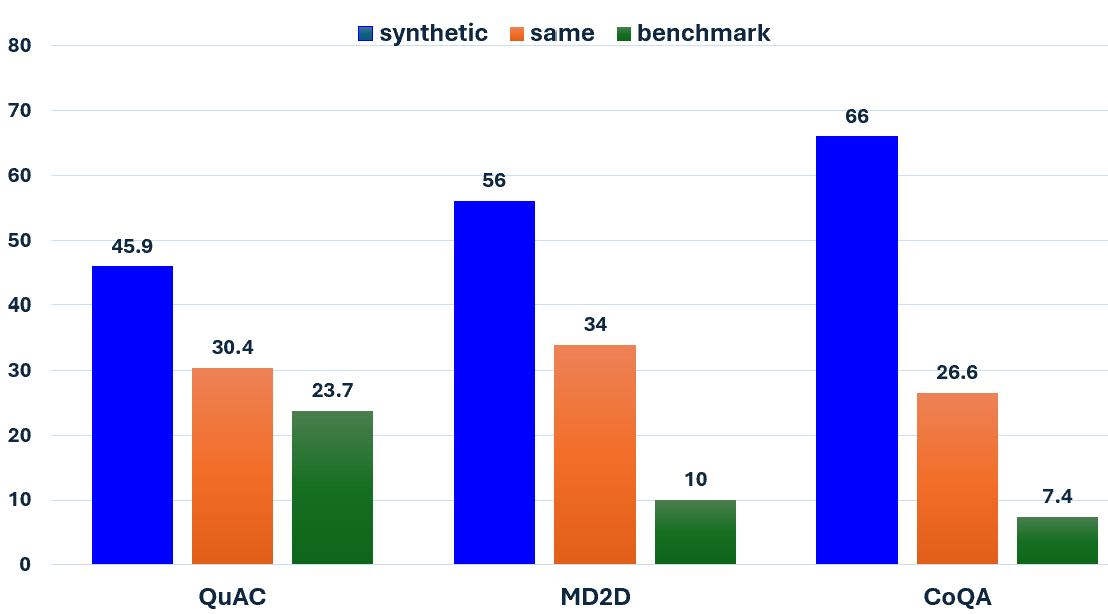}
    \caption{Winrate of two sets of fine-tuned models on the test sets judged by human annotators. 
    }
    \label{fig:winrate}
    \vspace{-0.4cm}
\end{figure}
We designed a human evaluation protocol that asks for human preference on two model outputs. Instructions and sample questions are given in Figure~\ref{fig:human-preference-coqa} (Appendix).
We randomly draw samples from answerable queries, 135 from QuAC, 100 from MultiDoc2Dial, and 100 from CoQA.\footnote{Since each grounding document in OR-QuAC is about 3 times longer than others, posing too much burden on the annotators, we have not included OR-QuAC for human evaluation.} We recruited 1 annotator for QuAC and MultiDoc2Dial, and 3 others for CoQA among professional annotators with 3 or more years of experience. 
The results are shown in Figure~\ref{fig:winrate}. 
Across all three test sets, human annotators highly prefer the outputs from the models fine-tuned on the synthetic data to those fine-tuned on the human benchmark data. Especially for CoQA, for which automatic evaluations were not decisive in determining the quality of the two models, three human annotators overwhelmingly prefer \textit{synthetic} (66\%) to \textit{human benchmark} (7.4\%). 

\subsection{Multi-Document Grounded Dialogs}
\begin{table*}
\small
\centering
\begin{tabular}{llcccc}\hline
\textbf{Models} & \textbf{Algorithms} & \textbf{F1} & \textbf{RougeL} & \textbf{Bert-Recall} & \textbf{Recall} \\\hline
LlaMA2-13B-Chat & Algorithm 1 & 0.283 & 0.282 & 0.294 & 0.333 \\
    & Algorithm 2 & \textbf{0.302} & \textbf{0.300} & \textbf{0.314} & \textbf{0.360} \\\hline
Merlinite-7B & Algorithm 1 & 0.272 & 0.268  & 0.308 & 0.372 \\
    & Algorithm 2 & \textbf{0.282}  & \textbf{0.278}  & \textbf{0.313} & \textbf{0.374} \\\hline
\end{tabular}
\caption{Performance comparison of OR-QuAC test set on two sets of fine-tuned models. One fine-tuned on data generated by Algorithm~\ref{alg:mrc-dialog-algorithm}, and the other on the data generated by Algorithm~\ref{alg:rag-dialog-algorithm}. 
}
\label{tab:cross-training}
\end{table*}
To the best of our knowledge, we present the first multi-document grounded multi-turn dialog generation pipeline that mimics RAG deployments for passage retrieval, \S{\ref{sec:rag-style}}. To evaluate the effectiveness of this technique, we evaluate the OR-QuAC test set on two fine-tuned models: One fine-tuned on the synthetic dialogs generated according to Algorithm 1, grounded on single documents, and the other fine-tuned on the synthetic data generated according to Algorithm 2, using a retriever to provide multiple documents to generate answers. 
The results in Table~\ref{tab:cross-training} indicate that the model fine-tuned on the synthetic data generated with the RAG setting out-performs the one fine-tuned on the synthetic data generated with single document.

\section{Related Work}
\label{sec:relatedwork}



Baize \cite{baize2023} propose a seed-driven self-chat technique for generating open-domain multi-turn user-agent conversations by enabling ChatGPT to engage in a conversation with itself, using Quora and Stack Overflow as seeds. 
UltraChat~\cite{ding2023enhancing} demonstrated the efficiency of scaling up LLM-based conversation generation. Unlike our work, it focuses on covering a wide range of topics, rather than multi-turn conversations. 


Recent work has also explored different forms of grounding when generating conversations -- grounding on personas~\cite{persona1,persona2}, topics~\cite{wsdm2024}, conversation summaries~\cite{gunasekara2021summary}, conversation instructions~\cite{mohapatra-etal-2021-simulated-chats} and single-document~\cite{arafat,wsdm2024}. \citet{chen2023exploring} generate conversations by grounding on retrieved knowledge graphs. In contrast we  focus our work on multi-document grounded dialogs. 

QuAC~\cite{quac2018} 
is a human generated multi-turn conversational benchmark. The data collection process resembles RAG setup with non-RAG characteristics for many initial queries. OR-QuAC \cite{orquac2020} tackles this problem by replacing initial questions in QuAC with their context-independent rewrites provided in CANARD dataset \cite{canard2019}.
Our RAG-style multi-turn dialog generation closely resembles OR-QuAC.
Nonetheless, OR-QuAC relies on the human generated dialogs from QuAC and some of the rewrites from CANARD. Furthermore, their grounding passages are based on human generated queries and answers. 
Multidoc2Dial \cite{multidoc2dial2021} is designed for information-seeking dialogs grounded on multiple documents. Aside from the fact that grounding documents  is human generated in MultiDoc2Dial and automatically generated by retrievers in our work, we automatically generate both the agent answers and user queries whose style and content are guided by query taxonomy and CoT-prompting. 

\section{Conclusion}

We presented a multi-turn synthetic dialog generation technique that incorporates a novel idea of taxonomy driven user query generation with CoT prompting, dialog generation from multi-documents updated by a retriever and an LLM-as-a-Judge module to select dialogs with correct answers. 
Human evaluations suggest that the synthetic data is diverse, coherent, and includes mostly correct answers. Human and automatic evaluations indicate models fine-tuned on our synthetic dialogs out-perform those fine-tuned on human generated benchmark data for answerable queries. Performance comparisons of OR-QuAC testset on models fine-tuned on single- versus multi-document grounded data suggest that multi-document grounded data generation is more effective for multi-document grounded tasks.

\newpage
\section{Limitations}

The method employs \textsc{mixtral-8x7b-instruct-v0.1} in the public domain. Therefore, the generated synthetic data can be susceptible to the limitations of the language model, particularly the biases inherent in the training data which may be harmful leading to synthetic data with hate, abuse and social stereotypes.

For multi-document grounded generation, there is no straightforward way of applying unanswerable query taxonomy to document retrieval, which always tries to retrieve relevant documents to a given user query. The query answerability may be determined only after examining whether or not the retrieved passages contain the correct answer to the query. This is unlike the single-document grounded data generation pipeline for which the grounding document is provided before query generation and therefore, one can steer an LM to generate a query whose answer cannot be found in the provided document. 
In addition, the benchmark test sets used in the evaluations have dataset specific categorical answers for unanswerable queries whereas the synthetically generated dialogs have often correct natural, contextual responses with no lexical overlap with the categorical reference answers. This may under-report any benefit from the use of our synthetic data for evaluation of unanswerable queries.  
We leave evaluation and generation of unanswerable queries in a multi-document grounded dialog generation setup to future work.

\bibliography{anthology,custom}
\newpage
\clearpage
\appendix

\begin{figure*}[!ht]
\centering
\small
\begin{minipage}[h]{1.05\linewidth}
\begin{tcolorbox}[colback=gray!5,colframe=olive!40!black]
\begin{verbatim}
[INST] 
You are a helpful assistant. 
Your task is to generate a question that can be answered using information from one or more given documents. 
The question must not be simple and require to read all the documents and then reason to answer. 
To generate a question that is answerable from a given set of documents, think step-by-step. 
1. Read all the documents and understand it. 
2. Identify one or more portions from documents to generate a question from. 
3. Generate a question from the identified text portions. 
4. The question could come from one or more of the following categories, 
    (a). requestiong a direct fact such as entity or number, 
    (b). requestiong a process such as bunch of steps to accomplish a task, 
    (c). requestiong a binary yes or no answer, 
    (d). requestiong to summarize a process or a policy, 
    (e). requesting for an explainaton of a policy. 
The input documents are enclosed between the tags <document> </document> whereas the question should be 
enclosed in the <question> </question> tags. 
Next, following the token <explanation>, you should explain step-by-step how you should look for the answer 
in the documents. The explanation should also include why certain pieces of information in the document is 
relevant to answer the question. Also reason whether the information in multiple sections of the document 
can be combined to identify the answer. End the explanation with the token </explanation>. 
Next, you need to generate the answer to the question, using the generated explanation and the information 
in documents following the token <answer>. The answer must be factually coherent to the documents and must 
not include any facts that are not found in the documents. If the answer cannot be inferred by the 
information in the documents your answer should indicate that. The answer should contain a short summary of 
reasoning. End the answer with the token </answer>. 
Next, you need to reason whether the answer and the explanation are consistent. Generate the reasoning for 
the consistency evaluation following the token <consistency>, and then provide a yes or no answer for 
consistency and end it with the token </consistency>. 
Then, you need to generate the evidences for the statements made in the answer, following the token 
<evidence>. Evidences must be extracted sentences from the documents, and must not contain any tokens that 
are not present in the document. Present the evidences in a numbered list. End the evidence with the token 
</evidence>.  
[/INST]
\end{verbatim}
\end{tcolorbox}
\end{minipage}
\caption{Chain-of-Thought prompt used for generating direct questions}
\label{fig:direct-prompt}
\small
\begin{minipage}[h]{1.05\linewidth}
\begin{tcolorbox}[colback=gray!5,colframe=olive!40!black]
\begin{verbatim}
[INST] 
You are a helpful assistant. 
Given one or more documents, your task is to generate a comparative question. A comparative question 
includes the comparison between two or more entities or phrases in the document. To generate a comparative 
query, think step-by-step. 
1. Read the entire document. 
2. Check to see if there are two entities or phrases for which similarities and differences can be assessed
according to certain criteria. 
3. If there are such entities or phrases, generate a comparative question, following a pattern such as, 
What is the difference between A and B?, How does A differ from B?, How does A compare to B?, Is A different 
from B?, What is better, A or B? 
4. If no entities or phrases can be compared with each other, generate 'no comparative queries can be 
generated' as the question. The input document is enclosed between the tags <document> </document> whereas 
the question should be enclosed in the <question> </question> tags.  
Next, you should explain whether the answer to the generated question is available in each document. 
The explanation should include why a certain piece of information in a document is relevant to answer the
question and why a certain document might not contain the answer. Also reason whether the information 
in multiple documents can be combined to identify the answer. Generate the explanation following the token 
<explanation>. End the explanation with the token </explanation>. 
Next, you need to generate the answer to the question, using the generated explanation and the information 
in documents following the token <answer>. The answer must be factually coherent to the document and must 
not include any facts that are not found in the documents. If the answer cannot be inferred by the 
information in the documents your answer should indicate that. Refer to facts in document using the URLs 
in the documents. The answer should contain a summary of reasoning. End the answer with the token </answer>. 
Next, you need to reason whether the answer and the explanation are consistent. Generate the reasoning 
for the consistency evaluation following the token <consistency>, and then provide a yes or no answer for 
consistency and end it with the token </consistency>. 
Then, you need to generate the evidences for the statements made in the answer, following the token 
<evidence>. Evidences must be extracted sentences from the document, and must not contain any tokens that 
are not present in the document. Present the evidences in a numbered list. End the evidence with the token 
</evidence>. 
[/INST]
\end{verbatim}
\end{tcolorbox}
\end{minipage}
\caption{Chain-of-Thought prompt used for generating comparative questions}
\label{fig:compative-prompt}
\vspace{-60pt}
\end{figure*}

\begin{figure*}[!ht]
\centering
\vspace{1cm}
\small
\begin{minipage}[h]{1.05\linewidth}
\begin{tcolorbox}[colback=gray!5,colframe=olive!40!black]
\begin{verbatim}
[INST] 
You are a helpful assistant. 
Your task is to generate a question that can be answered by aggregating information from one or more given 
documents. The question must not be simple and require to read all the documents, aggregate information 
and then reason to answer. To generate a question that is answerable by aggregating information from a 
given set of documents think step-by-step. 
1. Read all the documents and understand it. 
2. Identify two or more related portions from the documents to generate a question from. 
3. Generate a question that requires aggregating information from the identified text portions. 
4. The aggregation question could require following to generate the answer, 
    (a). accumulation of information from different sections of the documents, 
    (b). compare and contrast information from different sections of the documents, 
    (c). summarize information, 
    (d). reason about multiple options avaialble at different sections and provide an answer. 
The input documents are enclosed between the tags <document> </document> whereas the question should be 
enclosed in the <question> </question> tags. 
Next, following the token <explanation>, you should explain step by step how you should look for the answer 
in the documents. The explanation should also include why certain pieces of information in the document is 
relevant to answer the question. Also reason whether the information in multiple sections of the document 
can be combined to identify the answer. End the explanation with the token </explanation>. 
Next, you need to generate the answer to the question, using the generated explanation and the information 
in documents following the token <answer>. The answer must be factually coherent to the documents and must 
not include any facts that are not found in the documents. If the answer cannot be inferred by the 
information in the documents your answer should indicate that. Refer to facts in document using the URLs 
in the documents. The answer should contain a summary of reasoning. End the answer with the token </answer>. 
Next, you need to reason whether the answer and the explanation are consistent. Generate the reasoning for 
the consistency evaluation following the token <consistency>, and then provide a yes or no answer for 
consistency and end it with the token </consistency>. 
Then, you need to generate the evidences for the statements made in the answer, following the token 
<evidence>. Evidences must be extracted sentences from the documents, and must not contain any tokens that 
are not present in the document. Present the evidences in a numbered list. End the evidence with the token 
</evidence>.  
[/INST]
\end{verbatim}
\end{tcolorbox}
\end{minipage}
\caption{Chain-of-Thought prompt used for generating aggregate questions}
\label{fig:aggregate-prompt}
\small
\begin{minipage}[h]{1.05\linewidth}
\begin{tcolorbox}[colback=gray!5,colframe=olive!40!black]
\begin{verbatim}
[INST] 
You are a helpful assistant. 
Your task is to generate user question which is unanswerable using information in the given set of documents. 
However, the question must be distantly related to the information in the provided documents. To generate 
an unanswerable question, think step-by-step. 
1. Read the entire set of documents and understand it. 
2. Identify the domain and the topic of the document. 
3. Imagine a paragraph, that is related to the domain and the topic, but unlikely to be available in any 
neighboring documents. This imaginary paragraph may contain information about a related competing product or 
a company, imaginary related person or an event or incorrect irrelevant information about an entity in 
the documents. 
4. Generate a question on the content in the imagined paragraph. The input documents are enclosed between the 
tags <document> </document>, whereas the imaginary paragraph should be enclosed in the <paragraph> </paragraph>. 
The generated question should be enclosed in the <question> </question> tags. 
Next, you should explain whether the answer to the question is available in the documents. The explanation
should include why certain pieces of information in the document is relevant to answer the question. 
Generate the explanation following the token <explanation>. End the explanation with the token </explanation>. 
Next, generate the answer question, using the generated explanation and the information in documents 
following the token <answer>. If the question is unanswerable, the answer should include the explanation 
for unanswerability, and provide a summary of a related answer. End the answer with the token </answer>. 
Next, you need to reason whether the answer and the explanation are consistent. Generate the reasoning 
for the consistency evaluation following the token <consistency>, and then provide a yes or no answer for 
consistency and end it with the token </consistency>. 
Then, you need to generate the evidences for the statements made in the answer, following the token <evidence>. 
Evidences must be extracted sentences from the document, and must not contain any tokens that are not present 
in the document. Present the evidences in a numbered list. End the evidence with the token </evidence>.  
[/INST]
\end{verbatim}
\end{tcolorbox}
\end{minipage}
\caption{Chain-of-Thought prompt used for generating unanswerable questions}
\label{fig:unanswerable-prompt}
\vspace{-60pt}
\end{figure*}

\begin{figure*}[!ht]
\centering
\small
\begin{minipage}[h]{1.05\linewidth}
\begin{tcolorbox}[colback=gray!5,colframe=olive!40!black]
\begin{verbatim}
[INST] 
You are a helpful assistant. 
Given one or more documents and a user query, your task is to generate the agent response and a subsequent 
dialog between the user and the agent, which always includes a follow-up question by the user. A follow-up 
question is used to seek additional information related to a previous statement or question and typically 
start with 'what about', 'how about', 'what if'. To generate a dialog ending with a follow-up question 
by the user, think step-by-step. 
1. Read the documents. 
2. Identify the main topic or the key point being discussed. 
3. Start with the given user query and then generate an agent response verifiable from the given document. 
4. Identify a specific aspect of the main topic directly related to the previous user query, which has not 
been addressed in the agent response. 
5. Generate the followup user query on the specific aspect of the main topic. Then generate an agent 
response to the followup query. If the followup query is unanswerable from the given set of documents, 
indicate that in the agent response. 
User turns start with User: and the agent turns start with ASSISSTANT ANSWER. Generate at least 3 user 
turns and 3 agent turns, respectively, per document.
[/INST]
\end{verbatim}
\end{tcolorbox}
\end{minipage}
\caption{Chain-of-Thought prompt used for generating conversations with follow-up questions}
\label{fig:followup-prompt}
\small
\begin{minipage}[h]{1.05\linewidth}
\begin{tcolorbox}[colback=gray!5,colframe=olive!40!black]
\begin{verbatim}
[INST] 
You are a helpful assistant. 
Given one or more documents and a user query and the agent response, your task is to generate a dialog 
between the user and the agent, which always includes a clarification sentence by the user. A clarification 
question seeks to obtain more information  or to clarify something unclear. A clarifcation question 
typically start with 'what do you mean by', 'can you tell me more', 'so does that mean' while clarifying 
statement may begin with 'I meant', 'what I meant.' To generate a dialog ending with clarification by the 
user, think step-by-step. 
1. Read the entire document. 
2. Identify the main topic or the key point being discussed. 
3. Start with the given user query and the agent response. 
4. Identify a specific aspect of the main topic directly related to the previous user query for which the 
agent response is vague or unclear.  
5. Generate the clarification question on the specific aspect that that agent respons was not clear. 
Then generate an agent response to the clarification question. If the clarification question is
unanswerable from the given set of documents, indicate that in the agent response. 
User turns start with User: and the agent turns start with ASSISSTANT ANSWER. Generate at least 3 user 
turns and 3 agent turns, respectively, per document.
[/INST]
\end{verbatim}
\end{tcolorbox}
\end{minipage}
\caption{Chain-of-Thought prompt used for generating conversations with clarification questions}
\label{fig:clarification-prompt}
\small
\begin{minipage}[h]{1.05\linewidth}
\begin{tcolorbox}[colback=gray!5,colframe=olive!40!black]
\begin{verbatim}
[INST] 
You are a helpful assistant. 
Given one or more documents, your task is to generate a dialog between a user and an agent, which always 
ends with a correction statement from the user. A correction statement may start with 'no that's not what 
I meant', 'sorry, I mean'. To generate a correction statement by a user, think step-by-step. 
1. Read the entire document. 
2. Identify the main topic or the key point being discussed. 
3. Start with a user question leading to more than one possible agent responses. 
4. Identify the specific aspect of the previous user query that leads to multiple possible responses. 
5. Generate the correction statement that correct a previous user statement which prompts an alternative 
agent response. Then generate an agent response to the which takes into account the correction statement
made in the last turn. 
User turns start with User: and the agent turns start with ASSISSTANT ANSWER. Generate at least 3 user 
turns and 3 agent turns, respectively, per document.
[/INST]
\end{verbatim}
\end{tcolorbox}
\end{minipage}
\caption{Chain-of-Thought prompt used for generating conversations with correction questions}
\label{fig:correction-prompt}
\end{figure*}

\begin{figure*}
\small
\begin{minipage}[h]{1.05\linewidth}
\begin{tcolorbox}[colback=gray!5,colframe=olive!40!black]
\begin{verbatim}
[INST] 
You are a helpful and faithful assistant. Given one or more documents, a query and a response to the 
query, your task is to determine if the response to the query is correct or not. To determine the 
correctness of the response to the query, think step-by-step. 1. Check to see if the response can be 
decomposed into meaningful sub-responses such as multiple clauses. 2. If the response is atomic and 
cannot be decomposed into sub-responses, see if the response directly addresses the query. If not, 
the response is incorrect. If the response directly addresses the query, read the entire document and 
see if the response is verifiable from the document. If the response can be verified in the document, 
the response is correct. Otherwise, the response is incorrect. 3. If the response can be further 
decomposed into meaningful sub-responses, check to see if each of the sub-reponses directly addresses the 
query and can be verified in the document. If all of the sub-responses directly address the query and can be 
verified in the given document, the response is correct. Otherwise, the response is incorrect. Generate 
your answer either 'correct' or 'incorrect' starting with the token <answer> and ending with </answer>. 
Finally, generate step-by-step explanations justifying your answer starting with <explanation> and ending 
with </explanation>.
[/INST]
\end{verbatim}
\end{tcolorbox}
\end{minipage}
\caption{Chain-of-Thought prompt used for LLM-as-a-Judge, assessing answer correctness}
\label{fig:llm-judge}
\end{figure*}

\begin{table*}[ht]
\small
\centering
\begin{tabular}{l|l|l}\hline
\textbf{Evaluation Target} & \textbf{Criteria} & \textbf{Descriptions} \\\hline
Query & Answerability & Is the answer to the query present in the given documents? \\
      & Plausibility & How likely is the query created by an actual user? \\\hline
Response & Correctness & Is the response to the query correct on the basis of the information in the documents? \\\hline
Dialog & Diversity & Is each query-response in the dialog sufficiently different from each other? \\
       & Coherency & Is the dialog flow natural and coherent? \\\hline
\end{tabular}
\caption{Synthetic Data Quality Evaluation Criteria}
\label{tab:sdg-evaluation-criteria}
\end{table*}

\begin{figure*}[!ht]
\small
\begin{minipage}[h]{1.05\linewidth}
\begin{tcolorbox}[colback=gray!5,colframe=olive!40!black]
\begin{verbatim}
The primary purpose of the annotation is to assess the quality of synthetically generated user-agent 
conversations. Each dialog is grounded on one or more (up to five) document passages. Evaluation should be 
carried out on the basis of the information available from the documents according to the evaluation 
criteria. 

Note the relationship between query answerability and response correctness as follows:

- If the query is unanswerable, then the correct response is 'I do not know/No information is
  available in the given document
- If the query is answerable, then the correct response is factual information from the given
  document.

In determining the correctness of a response, follow the procedure specified below:

- Check to see if the query consists of a single clause or multiple clauses
- If the query consists of a single clause, then look for possible answers in the documents

  * If the possible answers overlap with the provided response, then the response is correct.
  * If the response is correct w.r.t. the query and contains some extra information,  the response is 
    deemed correct as long as the extra information does not contradict the main answer.

- If the query consists of multiple clauses, repeat 2 until all clauses in the query have been
  answered
\end{verbatim}
\end{tcolorbox}
\end{minipage}
\caption{Instructions for Synthetic Data Quality Evaluation}
\label{fig:sdg-evaluation-instruction}
\end{figure*}
\begin{table*}[!ht]
\small
    \centering
    \begin{tabular}{lccc}
        \hline
        Dataset        & Train Size & Dev Size & Test Size \\
        \hline
        CoQA           & 108,647   & 7,983    & -    \\
        QuAC           & 11,567    & 1,000    & -    \\
        MultiDoc2Dial  & 38,622   & 2,039    & 10,203    \\
        OR-QuAC        & 31,536    & 3,430    & 5,571    \\
        \hline
    \end{tabular}
    \caption{Data sizes for the train, dev, and test splits of CoQA, QuAC, MultiDoc2Dial, and OR-QuAC datasets.}
    \label{tab:data_sizes}
\end{table*}
\begin{table*}[ht]
\small
\centering
\begin{tabular}{lcccccc}\\\hline
\textbf{Benchmarks} & \textbf{Query Taxonomy} & \textbf{\#Dialogs} & \textbf{\#Turns/Dialog} & \textbf{Query Length} & \textbf{Response Length} & \textbf{Doc. Length} \\\hline
CoQA & aggregate & 9,967 & 6.2  & 14.7 & 41.6  & 268  \\
 & comparative & 8,894 & 5.5 & 12.6 & 35.1 & 270 \\
 & direct & 9,217 & 5.4 & 13.0  & 35.1  & 271 \\\hline
MD2D & aggregate & 8,572 & 5.0  & 21.6 & 54.5 & 635  \\
 & comparative & 8,211 & 5.1 & 15.5 & 55.4 & 720 \\
 & direct & 9,715 & 5.1 & 18.3  & 50.0  &  671 \\
 & unanswerable & 2,500  & 2  & 16.4 & 7.4 & 648  \\\hline
QuAC & aggregate & 9,271 & 6.0  & 16.1 & 47.1  & 461  \\
 & comparative & 9,554 & 6.4 & 12.6 & 44.8 & 474 \\
 & direct & 9,896 & 6.4 & 13.8  & 42.3  & 465 \\\hline
OR-QuAC & aggregate & 9,006 & 4.8  & 16.9 & 49.9  & 1,287 \\
 & comparative & 9,993 & 5.6 & 12.9  & 50.4  &  1,314 \\
 & direct & 5,000  & 5.6  & 14.4  & 48.0   & 1,134 \\\hline
(OR-)QuAC & unanswerable & 3,000  & 2  & 15.1 & 6.2 & 478  \\\hline

\end{tabular}
\caption{Multi-turn synthetic dialog statistics. Dialogs are generated with the training portion of the benchmark datasets as grounding documents according to Algorithm~\ref{alg:mrc-dialog-algorithm} for CoQA, MD2D and QuAC. For OR-QuAC, dialogs are generated according to Algorithm~\ref{alg:rag-dialog-algorithm} using the initial queries generated for QuAC as the only input. We generate unanswerable queries from QuAC documents and use them for both QuAC and OR-QuAC model training.}
\label{tab:sdg_stats}
\end{table*}



\begin{figure*}[!ht]
\small
\begin{minipage}[h]{1.05\linewidth}
\begin{tcolorbox}[colback=gray!5,colframe=olive!40!black]
\begin{verbatim}
Documents
Passage 0: The price-earnings ratio is calculated as the market value per share divided by the earnings per
share over the past 12 months. In your example, you state that the company earned $0.35 over the past 
quarter. That is insufficient to calculate the price-earnings ratio, and probably why the PE is just given 
as 20. So, if you have transcribed the formula correctly, the calculation given the numbers in your example 
would be: 0.35 * 4 * 20 = $28.00 As to CVRR, I'm not sure your PE is correct. According to Yahoo, the PE for 
CVRR is 3.92 at the time of writing, not 10.54. Using the formula above, this would lead to: 2.3 * 4 * 3.92 = 
$36.06. That stock has a 52-week high of $35.98, so $36.06 is not laughably unrealistic. I'm more than a 
little dubious of the validity of that formula, however, and urge you not to base your investing decisions 
on it.

Passage 1: PE ratio is the current share price divided by the prior 4 quarters earnings per share. Any stock 
quote site will report it. You can also compute it yourself. All you need is an income statement and a 
current stock quote.

Passage 2: 
...


Passage 3: 
...

Passage 4: The price to earnings ratio is a measure of the company's current share price compared to the 
annual net earnings per share. The other way to think about this is the number of years a company would take 
to pay back the share price if the earnings stay constant. This ignores factors like inflation and can be 
used as an indicator of risk. During the internet bubble many companies had P/E above 24 and no possible 
means of earning back the share prices that were inflated largely due to speculation. Most tools like 
Google Finance will list the P/E for a particular quote.

Dialog starts here:

user0: What might be the reasons that some firms don't report their Price-to-Earnings ratio and what could 
it imply when a company has a negative PE or no PE at all?
Answerability(required)
NO	YES	
Plausibility(required)
NO	YES	

agent0: Some firms may not report their Price-to-Earnings (PE) ratio because they have negative earnings,
and therefore the ratio cannot be calculated. A negative PE or no PE at all could imply that the company 
is not profitable or has had a one-time occurrence that affected its earnings, or it could be an awesome 
high growth company heavily investing for future profits. In the case of IPO companies, they often have 
negative earnings or no earnings reported as they are using the additional influx of cash from going public 
to grow the company.

Correctness(required)
NO	YES	

user1: Why might a company choose to not report their PE ratio even if they have positive earnings?
Answerability(required)
NO	YES	
Plausibility(required)
NO	YES	

agent1: I don't have enough information to provide a specific reason for every company, but in general, 
a company may choose not to report their PE ratio, even with positive earnings, if they have concerns about
the validity of the formula or if they don't want investors to base their decisions solely on the PE ratio, 
as it does not tell the whole story about a company's financial health.

Correctness(required)
NO	YES	

Dialog Level:
Diversity(required)
NO	YES	
Coherency(required)
NO	YES	
\end{verbatim}
\end{tcolorbox}
\end{minipage}
\caption{Sample Synthetic Data Quality Evaluation}
\label{fig:sdg-quality-evaluation-sample}
\vspace{-60pt}
\end{figure*}

\begin{figure*}[!ht]
\small
\begin{minipage}[h]{1.05\linewidth}
\begin{tcolorbox}[colback=gray!5,colframe=olive!40!black]
\begin{verbatim}
Instruction

Given one or more grounding documents, dialog history, user query and 2 agent responses, your task is to 
select your preferred agent response. Please read the entire grounding document carefully before choosing 
one of the given options

Document

CHAPTER XVII. DEFEAT. A grey dimness was creeping in upon the schooner when a little bitter breeze sprang 
up from westwards, and Dampier bade them get the mainsail on to her. "I don't like the look of the weather, 
and I'm beginning to feel that I'd like to see that boat," he said. "Anyhow, we'll get way on her." It was a 
relief to hoist the mainsail. The work put a little warmth into them, and the white men, at least, had been 
conscious of a growing uneasiness about their comrades in the boat. The breeze had, however, freshened 
before they set it, and there were white caps on the water when the _Selache_ headed for the ice. It had 
somewhat changed its formation when they approached it, for big masses had become detached from it and were 
moving out into the open water, while the opening had become perceptibly narrower. The light was now fading 
rapidly, and Wyllard took the wheel when Dampier sent the man there forward. "Get the cover off the second 
boat, and see everything clear for hoisting out," he said to him, and then called to Wyllard. "We're close 
enough. You'd better heave her round." She came round with a thrashing of canvas, stretched out seawards, 
and came back again with her deck sharply slanted and little puffs of bitter spray blowing over her weather 
rail, for there was no doubt that the breeze was freshening fast. Then Dampier sent a man up into the 
foremast shrouds, and looked at Wyllard afterwards.

Context

User: What was heading towards the boat?
Agent: a grey dimness
User: What came from the west?
Agent: a bitter breeze
User: Who gave some orders?
Agent: Dampier
User: What did he want done?
Agent: get the mainsail on
User: Did he feel good about the weather?
Agent: no
User: What did he want to see?
Agent: that boat
User: What did they put up?
Agent: the mainsail

Query: User: How did doing this make them feel?

Responses
Agent1: it put a little warmth into them
Agent2: a little warmth

Which response is better(required)
 Agent1
 Agent2
 Both correct
 Both incorrect

\end{verbatim}
\end{tcolorbox}
\end{minipage}
\caption{Human Preference Evaluation Sample from CoQA Answerable Query}
\label{fig:human-preference-coqa}
\end{figure*}

\begin{figure*}[!ht]
\small
\begin{minipage}[h]{1.05\linewidth}
\begin{tcolorbox}[colback=gray!5,colframe=olive!40!black]
\begin{verbatim}
Document:

"On Sundays my father always wore that dull gray apron - the one with the race cars all over it. The ritual 
began after breakfast when Dad always announced: "Go ahead everyone. I'll take care of the dishes!" 
With that my mother disappeared into the folds of the Sunday paper. Off came the suit coat he had worn to 
church that morning. Up went the shirtsleeves. On went that apron. For the next hour Dad did the dishes, 
singing ballads like "I Had a Hat When I Came In" and "Who Put the Chow in Mrs. Murphy's Chowder?"
I suppose it was strange for a boy's father to wear an apron - even one with race cars - but I never thought
much of it until the day that Dad broke with tradition. It was the last Sunday in August. My father seemed 
in an expansive mood as we walked home from church together. 
"Tommy," he said letting my name roll off his tongue. My mind raced ahead of his words: The birds and the 
bees? A new bike? A part-time job? "There comes a time in every boy's life when he must take on
responsibilities." This was important. I might even get to back the car out of the driveway.
"Responsibilities?" I asked. "Yes. It's time you took a greater role in the household." Power tools? 
Boss my baby brother? "Starting today, I want you to do the dishes on Sunday morning so your mother and I can 
work the crossword puzzle together." "The dishes!?" "Anything wrong with taking over the dishes, son?" 
I started to say something about a man's job or woman's work, but I knew immediately that my protests would 
fall on deaf ears. I didn't taste a bit of breakfast that morning. Dad seemed in a jovial mood as he 
described an exceptional Yankee game seen through the eyes of Mel Allen on the radio last night. 
"Mickey Mantle drove the ball right over the center field wall," he said. 
"Just a straight line climb in right out of the stadium." He looked out the window as if trying to pick the 
ball out of the cloud formations. I tried to imagine Mickey Mantle wearing an apron. Suddenly, everything 
grew quiet. My sister began to clear the table. My brother was scraping the last of the egg from his 
plate. And then that ancient family ritual that had filled so many Sunday mornings came to an end. My father 
announced: "Let's go read the paper, Hon." "Aren't you doing the dishes?" my mother asked in puzzlement. 
"Your oldest son has generously offered to fill the position." My brother and sister stopped cold. 
So this was what my life had come to. A dark angel sat on my left shoulder and reminded me that I 
could hit a baseball farther than anyone in my class. I could bench-press my weight. I knew three 
declensions in Latin, the language of Caesar. Ask me to run through a rainstorm. Command me to ride the 
roller coaster - backward. These things I would do. But I could never do those dishes. There was nothing 
left but to refuse. People often say there is a special chemistry between a father and a son. He came back 
into the kitchen just as I was about to storm out. He had loosened his tie and rolled up the sleeves of his 
shirt - ready to relax. In his right hand was the old apron. "I want you to have this, Tommy. 
It'll keep your clothes from getting wet." And before I could mount a protest, he had put the 
thing on me. "Thanks, Son. Your mother and I appreciate this." With that he disappeared into the Sunday 
paper. I looked down at the plastic. It had seen better days. I could see my dad reaching for the dishes. 
The dark angel flew off. Soon I was singing about Mrs. Murphy's chowder. 
The words came out of nowhere. And out of nowhere I knew the kind of man I wanted to be."

Conversation:

User: How did the father's behavior change towards the dishes on Sundays and why?
Agent: The father, who used to do the dishes every Sunday morning while singing ballads, passed on the 
          responsibility to his son, Tommy, starting from the last Sunday in August. He did this to give Tommy 
       a chance to take on responsibilities and play a greater role in the household.
User: What did the father give to Tommy on the day he passed on his responsibilities?
Agent: The father gave Tommy his old apron, which had race cars on it, to keep Tommy's clothes from getting 
       wet while doing the dishes.
User: What did the father do after giving the apron to Tommy?
Agent: After giving the apron to Tommy, the father disappeared into the Sunday paper, leaving the dishes 
      for Tommy to do. 


\end{verbatim}
\end{tcolorbox}
\end{minipage}
\caption{A sample synthetically generated conversation from CoQA documents}
\label{fig:coqa-example-conversation}
\end{figure*}

\begin{figure*}[!ht]
\small
\begin{minipage}[h]{1.05\linewidth}
\begin{tcolorbox}[colback=gray!5,colframe=olive!40!black]
\begin{verbatim}
Document:

Benefits Planner: Retirement 
If You Are Divorced 
If you are divorced, but your marriage lasted 10 years or longer , you can receive benefits on your 
ex - spouse's record even if they have remarried if : You are unmarried ; You are age 62 or older ; 
Your ex - spouse is entitled to Social Security retirement or disability benefits ; and The benefit you 
are entitled to receive based on your own work is less than the benefit you would receive based on your 
ex - spouse's work. If you remarry, you generally cannot collect benefits on your former spouse's record 
unless your later marriage ends whether by death, divorce, or annulment. If your ex - spouse has not 
applied for retirement benefits, but can qualify for them, you can receive benefits on their record if you 
have been divorced for at least two years. If you are eligible for retirement benefits on your own record 
and divorced spouse's benefits , we will pay the retirement benefit first. If the benefit on your 
ex - spouse's record is higher , you will get an additional amount on your ex - spouse's record so that 
the combination of benefits equals that higher amount. If you continue to work while receiving benefits , 
the retirement benefit earnings limit still applies. If you are eligible for benefits this year and are 
still working , you can use our earnings test calculator to see how those earnings would affect your 
benefit payments. If you will also receive a pension based on work not covered by Social Security, such as 
government work , your Social Security benefit on your ex - spouse's record may be affected. 

Conversation:

User: Can a person who is divorced and currently married receive benefits on their ex-spouse's record
Agent: No, if you remarry, you generally cannot collect benefits on your former spouse's record unless your 
       later marriage ends whether by death, divorce, or annulment.
User: What happens if the ex-spouse has not applied for retirement benefits but can qualify for them?
Agent: If your ex-spouse has not applied for retirement benefits, but can qualify for them, you can receive 
       benefits on their record if you have been divorced for at least two years.
User: So does that mean I can only receive benefits on my ex-spouse's record after two years of being 
      divorced, even if they haven't applied for benefits yet?",
Agent: Yes, that's correct. You can receive benefits on their record if you have been divorced for at least 
      two years.
User: What if I am also eligible for retirement benefits?
Agent: In that case, if your ex-spouses's record is higher, you will receive an additional amount so that 
       the combination of benefits equals that higher amount. 


\end{verbatim}
\end{tcolorbox}
\end{minipage}
\caption{A sample synthetically generated conversation from MultiDoc2Dial documents}
\label{fig:md2d-example-conversation}
\end{figure*}


\begin{table*}[ht]
\small
\centering
\begin{tabular}{llcccc}\hline
\textbf{Benchmarks/Models}  & \textbf{TrainSet} & \textbf{F1} & \textbf{RougeL} & \textbf{Bert-Recall} & \textbf{Recall} \\\hline
\textbf{QuAC}  (335)          \\\hline
Mixtral-8X7B-Instruct-v0.1 & zero-shot & 0.139 & 0.166 & 0.150 & 0.451 \\\hline
LlaMA2-13B-Chat   & zero-shot & 0.203 & 0.235 & 0.092 & 0.318 \\
         & human benchmark & 0.843   & 0.852  & 0.805  & 0.844  \\
         & synthetic & 0.527  & 0.550  & 0.440  & 0.540 \\
         & synthetic-filtered & 0.431 & 0.456 & 0.340 & 0.453
         \\\hline
Merlinite-7B  & zero-shot &  0.115 & 0.143 & 0.044 & 0.313 \\   
     & human benchmark & 0.879 & 0.888 & 0.852 & 0.881 \\
     & synthetic & 0.467 & 0.494 & 0.366 & 0.479 \\
     & synthetic-filtered & 0.258 & 0.292 & 0.130 & 0.322 \\
\hline
\textbf{OR-QuAC} (1150)           \\\hline
Mixtral-8X7B-Instruct-v0.1 & zero-shot & 0.013 & 0.012 & 0.030  &  0.048 \\\hline
LlaMA2-13B-Chat & zero-shot & 0.010   & 0.009  & 0.022 & 0.037   \\
  & human benchmark & 0.799   & 0.799  & 0.799  & 0.799  \\
    & synthetic & 0.468  &  0.465  & 0.469  & 0.472 \\
    & synthetic-filtered & 0.036 & 0.036 & 0.036 & 0.036 \\
\hline
Merlinite-7B & zero-shot & 0.025 & 0.024 & 0.023 & 0.027 \\
   & human benchmark & 0.841  & 0.841 & 0.841 & 0.841  \\
   & synthetic & 0.215  & 0.215 & 0.216 & 0.217  \\
   & synthetic-filtered & 0.011 & 0.011 & 0.012 & 0.013 \\
\hline
\textbf{MD2D} (4026)           \\\hline
Mixtral-8X7B-Instruct-v0.1 & zero-shot & 0.118 & 0.165 & 0.191 & 0.255 \\\hline
LlaMA2-13B-Chat & zero-shot & 0.179 & 0.193 & 0.217 & 0.289 \\
    & human benchmark &  0.766 &	0.753 &	0.779 &	0.755  \\
    & synthetic &  0.461 & 0.441 &	0.580 &	0.611 \\
    & synthetic-filtered &  0.459 & 0.439 &	0.581 &	0.613 \\
\hline
Merlinite-7B & zero-shot & 0.181  & 0.199  & 0.277 & 0.290 \\
   & human benchmark &  0.769 &	0.756 &	0.787 &	0.765  \\
   & synthetic &  0.410 &	0.403 &	0.546 &	0.585  \\
   & synthetic-filtered &  0.411 &	0.403 &	0.536 &	0.543  \\
\hline
\end{tabular}
\caption{Benchmark test set performance on unanswerable queries. Comparison is between models fine-tuned on human annotated benchmark training data, \textit{human benchmark}, synthetic data without LLM-as-a-Judge filter (and with single turn unanswerable queries), \textit{synthetic}, and synthetic data with LLM-as-a-Judge filter (and without single turn unanswerable queries), \textit{synthetic-filtered}. For CoQA, unanswerable queries take up less than 1\% of the benchmark test data (65 samples) and therefore, we do not include the results for CoQA in this table. Models fine-tuned on synthetic data (synthetic) under-perform those fine-tuned on human benchmark training data (human benchmark), which may be partially attributed to the fact that the benchmark test sets used in the evaluations have dataset specific categorical answers for unanswerable queries whereas the synthetically generated dialogs have often correct natural, contextual responses with no lexical overlap with the categorical reference answers. The performance gap between synthetic (trained with unanswerable queries) and synthetic-filtered (trained without unanswerable queries) indicate that synthetically generated single-turn unanswerable queries are effective for multi-turn dialogs.}
\label{tab:only-unanswerable-results}
\end{table*}

\end{document}